\documentclass[letterpaper]{article} 

\ifdefined\aaaianonymous
    \usepackage[submission]{aaai2026}  
\else
    \usepackage{aaai2026}              
\fi

\usepackage{times}  
\usepackage{helvet}  
\usepackage{courier}  
\usepackage[hyphens]{url}  
\usepackage{graphicx} 
\usepackage{natbib}  
\usepackage{caption} 
\usepackage{amsmath}    
\usepackage{booktabs}   
\usepackage{tabularx}   
\usepackage{array}          
\newcolumntype{C}[1]{>{\centering\arraybackslash}m{#1}} 
\newcolumntype{Y}{>{\centering\arraybackslash}X}        
\usepackage{placeins}  
\usepackage{booktabs}  
\usepackage{tabularx}  
\usepackage{multirow}   
\usepackage{siunitx}    
\usepackage{subcaption}  
\newcolumntype{M}[1]{>{\raggedright\arraybackslash}m{#1}}

\newif\ifarxiv

\usepackage{algorithm}
\usepackage{algorithmic}
\usepackage{pifont}

\usepackage{newfloat}
\usepackage{listings}
\DeclareCaptionStyle{ruled}{labelfont=normalfont,labelsep=colon,strut=off} 
\floatstyle{ruled}
\newfloat{listing}{tb}{lst}{}
\floatname{listing}{Listing}

\ifdefined\aaaianonymous
     \title{Advancing Multimodal Teacher Sentiment Analysis: \\ 
The Large-Scale T-MED Dataset \& The Effective AAM-TSA Model}
 
\else
    \title{Advancing Multimodal Teacher Sentiment Analysis: \\ 
The Large-Scale T-MED Dataset \& The Effective AAM-TSA Model}
\fi

\author{
    Zhiyi Duan\textsuperscript{\rm 1},
    Xiangren Wang\textsuperscript{\rm 1},
    Hongyu Yuan\textsuperscript{\rm 1},
    Qianli Xing\textsuperscript{\rm 2}\thanks{Corresponding author.}
}

\affiliations {
    \textsuperscript{\rm 1}Department of Computer Science, Inner Mongolia University, Hohhot, Inner Mongolia\\
    \textsuperscript{\rm 2}College of Computer Science and Technology, Jilin University, Changchun, Jilin\\
    duanzy@imu.edu.cn, wangxr@mail.imu.edu.cn, yuanhongyu\_1997@163.com, qianlixing@jlu.edu.cn
}

\usepackage{bibentry}

\begin{document}

\maketitle

\begin{abstract}
Teachers' emotional states are critical in educational scenarios, profoundly impacting teaching efficacy, student engagement, and learning achievements. 
However, existing studies often fail to accurately capture teachers'  emotions due to the performative nature and overlook the critical impact of instructional information on emotional expression.
In this paper, we systematically investigate teacher sentiment analysis by building both the dataset and the model accordingly. We construct the first large-scale teacher multimodal sentiment analysis dataset, T-MED.
To ensure labeling accuracy and efficiency, we employ a human-machine collaborative labeling process.
The T-MED dataset includes 14,938 instances of teacher emotional data from 250 real classrooms across 11 subjects ranging from K-12 to higher education, integrating multimodal text, audio, video, and instructional information.
Furthermore, we propose a novel asymmetric attention-based multimodal teacher sentiment analysis model, AAM-TSA.
AAM-TSA introduces an asymmetric attention mechanism and hierarchical gating unit to enable differentiated cross-modal feature fusion and precise emotional classification. Experimental results demonstrate that AAM-TSA  significantly outperforms existing state-of-the-art methods in terms of accuracy and interpretability on the T-MED dataset.

\end{abstract}
\section{Introduction}
In the intricate scenarios of modern education, the emotional states of teachers emerge as a pivotal factor, profoundly influencing not only the efficacy of pedagogical delivery but also the crucial aspects of student engagement, motivation, and ultimately, learning achievements \cite{HAN2023103138}. Consequently, the systematic analysis of teacher emotions has garnered increasing attention within educational psychology and human-computer interaction research \cite{WANG2025101456}. 

Unlike general sentiment analysis, analyzing teacher emotions is distinguished by several unique characteristics. On the one hand, the performative nature of teaching requires educators to project specific emotions, such as enthusiasm or composure. Thus, teachers frequently exhibit hidden emotions, deliberately suppressing feelings like frustration or stress to uphold professionalism or shield students from negative influences\cite{KingAlmukhaildMercerBabicMairitschSulis+2024+1213+1235}. On the other hand, teacher emotions are inherently affected by instructional information, including instructional subject and educational stage\cite{HARGREAVES2000811}.

The current approaches to teachers' sentiment analysis have predominantly relied on single-modal data modeling, primarily focusing on textual transcripts or isolated audio cues \cite{2024A}. 
While these methods have offered valuable preliminary insights, their inherent limitations become apparent when confronted with the complexity and richness of real-world teaching environments. 
The inadequacy of single-modal data is further amplified by the distinctive nature of teachers' emotional expressions, characterized by deliberate deductiveness, e.g., emotive cues tailored to instructional goals, and strategic restraint, e.g., suppression of negative emotions to maintain classroom dynamics. In summary, rendering any single data modality is insufficient to disentangle performative displays from underlying emotion states.

Teacher sentiment analysis is inherently multimodal, manifested through a confluence of text, audio, video, and instructional information \cite{cheng2024emotionllama}. 
This integration can effectively capture environment-induced variations in emotional expression and decode implicit affective cues, which are critical for understanding teachers' true emotional experiences. 
Although current multimodal models try to integrate information from different modalities \cite{2024A,10517669}, they often fail to effectively analyze teachers' emotions due to the performative nature. On the other hand, the existing multimodal datasets \cite{7961761, bagher-zadeh-etal-2018-multimodal} are mostly designed for general sentiment analysis, which further interferes with the performance of the sentiment analysis models.
Thus, there is an urgent need for both a high-quality multimodal dataset and specialized multimodal teacher sentiment analysis models.

To bridge this critical gap and facilitate a more holistic and accurate understanding of teacher emotional dynamics, this paper presents a systematic investigation into teacher sentiment analysis, encompassing both data construction and model development. 
We first present a novel human-machine collaboration annotation collaborative framework, and collect a large-scale dataset named T-MED, containing four modalities. We specially designed eight sentiment labels, including three types of teacher-specific emotions: patience, enthusiasm, and expectation \cite{Zhang2024Guiding, Frenzel2018Emotion,RubieDavies2023Beyond}.
Then, we propose a novel asymmetric attention-based multimodal teacher sentiment analysis model, AAM-TSA. We have crafted an audio-focused asymmetric attention mechanism to seamlessly unify multimodal features into a cohesive representation, while maintaining unique modality-specific traits and capturing inter-modal semantic relationships.
In summary, our contributions are as follows:
\begin{itemize}
\item   We construct a large-scale teacher multimodal sentiment analysis dataset, T-MED, for the first time.
The T-MED dataset contains 14,938 samples of 4 modalities across 8 types of teacher emotional data, spanning over 17 hours in total and encompassing multidisciplinary scenarios from K-12 to higher education.
\item We propose a novel asymmetric attention-based multimodal teacher sentiment analysis model, AAM-TSA. From the perspectives of multimodal information learning and fusion, AAM-TSA deeply alleviates the deductive problem in teacher's emotional analysis.
\item The experimental results show that the collected T-MED dataset and the proposed  AAM-TSA are effective and reasonable. AAM-TSA is superior to the state-of-the-art multimodal sentiment analysis methods on the T-MED.
\end{itemize}

\section{Relate Work}

\subsection{Teachers' Sentiment Analysis}
Teacher sentiment analysis has undergone a significant evolution, transitioning from single-modal to multimodal approaches in response to the growing demand for a more comprehensive understanding of emotional expression in educational contexts \cite{li2022emotion,shaikh2019aspects,lu2024review}. For instance, Zhang et al. extracted acoustic spectrum features, including Mel-frequency cepstral coefficient (MFCC) features and filter bank (FBank) features, utilizing parallel subnet training structures for feature fusion \cite{zhang2022research}.

Consequently, there has been a notable shift towards multimodal frameworks that integrate data from multiple sources.
ProsodyBERT designed a parallel coding architecture to process pronunciation, MFCC and text features  \cite{2024A}.
AMTBA proposed a classroom event segmentation method based on voice activity detection (VAD), emphasizing the structured information processing of classrooms \cite{10517669}. 

Despite these advancements, current multimodal methods are constrained by several critical limitations. A primary challenge lies in the effective fusion of heterogeneous data types, which necessitates sophisticated models capable of managing the complexity and variability inherent in multimodal inputs. Moreover, the development and evaluation of these models are significantly hindered by the scarcity of large-scale, annotated multimodal datasets, which are essential for training robust teachers' sentiment analysis models.

\subsection{Datasets for Teachers' Sentiment Analysis}
It is necessary to construct the domain-specific teacher sentiment analysis datasets as the general multimodal datasets for sentiment analysis \cite{7961761, bagher-zadeh-etal-2018-multimodal}  are hard to apply to teachers' sentiment analysis due to the performative nature of teachers.
He et al. constructed 1,825 pure audio clips, directly pointing to the core modality of teacher sentiment analysis \cite{10661152}.
AMTBA dataset contained 2,109 voice emotion data for classroom voice emotion performance testing\cite{10517669}. 
TESD dataset was constructed with a total of 3564 emotional voice samples from 6 male and 6 female teachers,  which includes four emotional categories: anger, surprise, neutrality, and sadness\cite{10.1007/978-981-97-0730-0_19}. 
These teacher sentiment analysis datasets often face limitations such as small scale, limited modality, and restricted access due to privacy concerns, which hinder their broader research and application.

\section{Human-Machine Collaboration Annotation Framework for T-MED Dataset}

\begin{figure*}
    \centering
    \includegraphics[width=\textwidth]{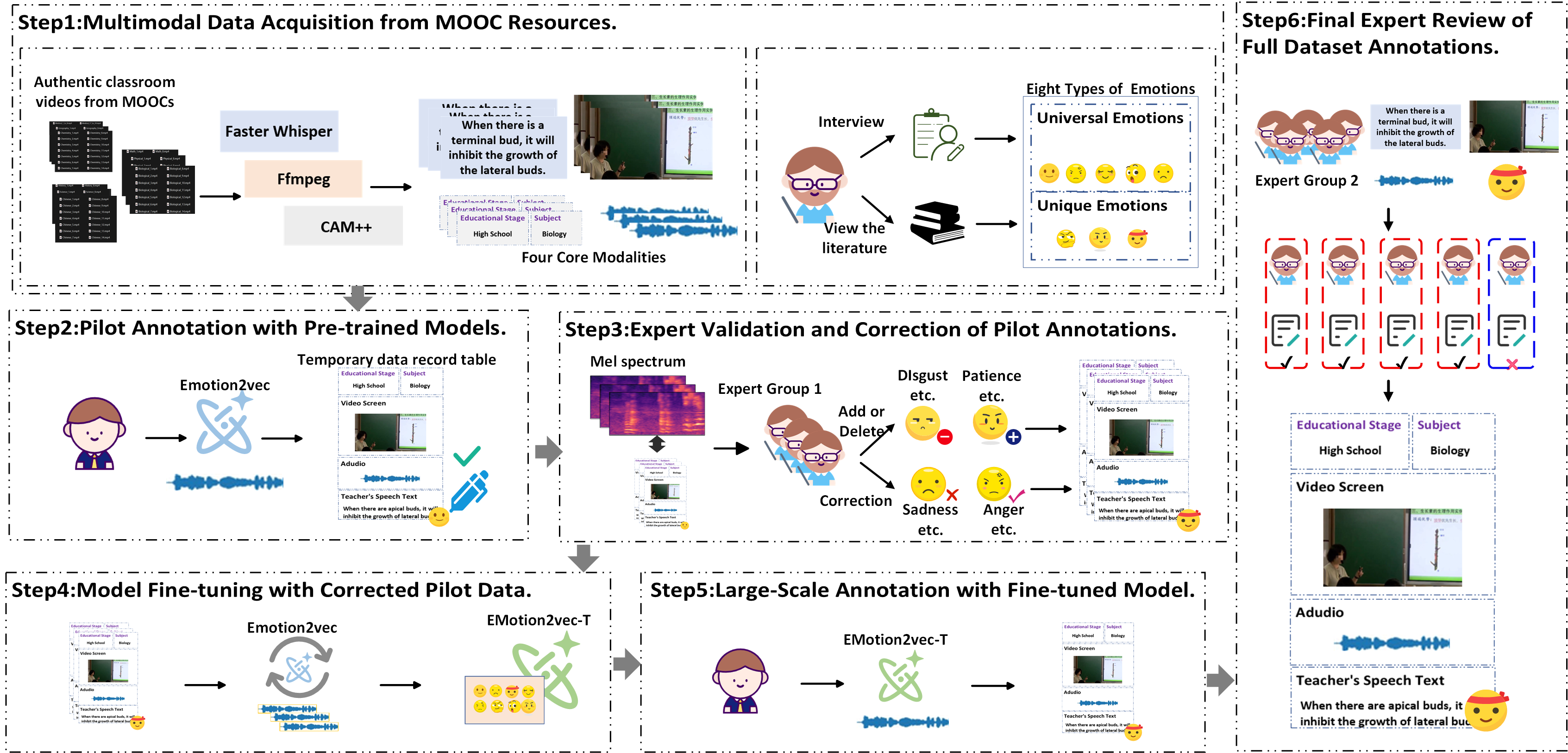}
    \caption{Diagram of T-MED dataset construction process.}
    \label{1}
\end{figure*}

To address the critical need for high-quality datasets in teachers' sentiment analysis, we present a novel human-machine collaboration annotation collaborative framework,  yielding a high-quality dataset named T-MED with reliable sentiment labels.
Our human-LLM annotation collaborative framework integrates iterative model refinement and expert oversight across six sequential steps as shown in Figure \ref{1}. A comprehensive explanation of each step is provided below.
Note that, we specially annotate three distinct sentiment categories: patience, enthusiasm, and expectation to capture the characteristics of teacher emotions. The detailed information of T-MED is provided in Appendix A.

\textbf{Step 1: Multimodal Data Acquisition from MOOC Resources.}
A total of 250 high-quality videos encompassing educational scenarios ranging from K-12 to higher education, are selected and downloaded from public MOOC platforms from multiple providers. 
To extract teacher-centric multimodal cues, open-source tools are employed for systematic processing: FFmpeg \cite{ffmpeg} for segmenting and preprocessing raw video files, Faster Whisper \cite{fasterwhisper} for generating accurate text transcripts from speech, and the Speaker Identification Model CAM++ \cite{clearervoice} for isolating teacher-specific audio streams. 

Finally, we obtain four core modalities: video clips, audio tracks, textual transcripts, and instructional information. Specifically, instructional information is a text-based modality encompassing two key dimensions: subject information and grade-level information, which are vital for contextualizing emotional expressions.
Note that, eight types of emotional labels are finally identified: neutral, anger, joy, surprise, sadness, patience, enthusiasm, and expectation. The former five sentiments are common emotions in the general field, while patience, enthusiasm, and expectation are more reflective of professional characteristics specific to teachers.

\textbf{Step 2: Pilot Annotation with Pre-trained Models.} 
A subset of the multimodal data (10\% of the total corpus) is subjected to initial emotion labeling via a pre-trained model emotion2vec-large \cite{ma-etal-2024-emotion2vec}. 
The purpose of this step is to generate preliminary emotional labels for subsequent expert review and model refinement while reducing the manual annotation burden for the larger corpus.

\textbf{Step 3: Expert Validation and Correction of Pilot Annotations.} 
A group of three experts is enlisted to conduct manual validation of the pre-labeling results generated in step 2. The experts collaboratively analyzed the initial pre-labeling outputs by cross-referencing audio clips with their corresponding Mel spectrograms. Using eight predefined emotional labels as guidance, they methodically refine inaccurate annotations and create a high-quality seed dataset for fine-tuning.

\textbf{Step 4: Model Fine-tuning with Corrected Pilot Data.} 
The pre-trained model is fine-tuned using the expert-corrected seed dataset, adapting its parameters to the unique emotional patterns of teachers. This refinement enhances the model’s ability to generalize to the broader dataset, minimizing domain-specific errors.

\textbf{Step 5: Large-Scale Annotation with Fine-tuned Model.} 
The entire multimodal dataset undergoes automated labeling using the fine-tuned model, leveraging its improved accuracy to process large volumes efficiently. This step scales the annotation workflow while maintaining consistency across the corpus.

\textbf{Step 6: Final Expert Review of Full Dataset Annotations.} 
A panel of five experts who are not involved in any prior annotation stages is formed to conduct the final review. These experts independently examine the predicted labels by integrating multimodal information. Adhering strictly to predefined consensus criteria, each annotation was validated through a rigorous voting mechanism: the data is included in the final dataset only when no fewer than four experts unanimously confirmed the correctness of a label; otherwise, the data is discarded.

\section{ Asymmetric Attention-based Multimodal Teacher Sentiment Analysis Model}
\begin{figure*}
    \centering
    \includegraphics[width=\textwidth]{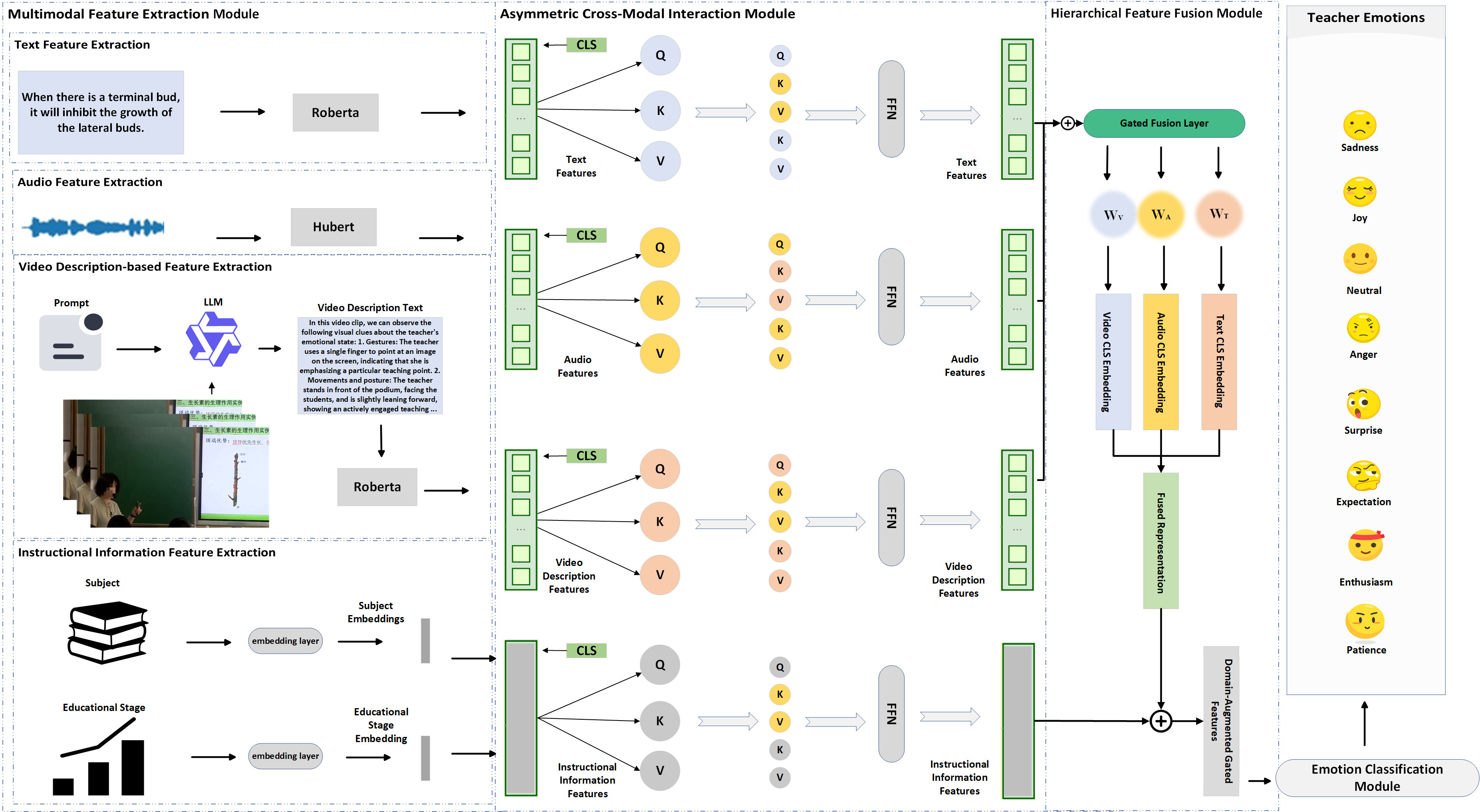}
    \caption{AAM-TSA model architecture diagram.}
    \label{2}
\end{figure*}

In this section, we propose a novel Asymmetric Attention-based Multimodal Teacher Sentiment Analysis Model, shorted as AAM-TSA, which consists of four modules to comprehensively utilize the information from four modalities, as shown in Figure \ref{2}.

We specially design an audio-centric asymmetric attention mechanism to integrate the multimodal features into a unified representation, preserving both modality-specific characteristics and inter-modal semantic associations.
The detailed descriptions of the four modules in AAM-TSA are provided in the following.

\subsection{Multimodal Feature Extraction Module}
Upon the T-MED dataset, we focus on processing four modalities of data: textual transcripts, audio tracks, video clips, and instructional information, all of which are crucial for mining the hidden emotional cues of teachers. 

We extract the corresponding feature vectors for each modality.
 For textual transcripts, we utilize RoBERTa\cite{liu2020roberta} for processing as shown in E.q. \ref{textual transcripts}, where \(h_t\) is the text feature matrix, and $T_{input}$  represents input token embeddings.
 \begin{equation}
 \label{textual transcripts}
h_t = \text{RoBERTa}(T_{input})
\end{equation}
%
For audio tracks, HuBERT\cite{9585401} is employed as shown in E.q. \ref{audio tracks}, where \(h_a\) is the audio feature matrix and $A_{input}$ is the raw waveform.
\begin{equation}
\label{audio tracks}
h_a = \text{HuBERT}(A_{input})
\end{equation}


Furthermore, we utilize a large language model to create text descriptions for video clips, using tailored prompts that emphasize four emotional dimensions: gestures, movements, eye contact, and facial expressions. The feature of video clips are extracted as follows:
\begin{equation}
V_{text}=\text{Video LLM}(V_{input}, P_{emo}),
\end{equation}
\begin{equation}
\quad h_v = \text{RoBERTa}(V_{text}), 
\end{equation}
where $V_{input}$ is the video clip; $P_{emo}$ is the emotion-focused prompt template,  and its specific content is placed  in the Appendix B. $V_{text}$ is the generated description text.\(h_v\) is the video feature matrix used in our scheme.

To effectively integrate instructional information, we convert discrete educational stage and subject labels into continuous representations to a unified 768-dimensional vector space through a fully connected layer, which is consistent with the dimension of text, audio, and video features. The feature of instructional information are extracted as follows:
\begin{equation}
\text{e}_l = \text{Embedding}_{\text{stage}}(l),
\end{equation}
\begin{equation}
\text{e}_s = \text{Embedding}_{\text{subject}}(s),
\end{equation}
\begin{equation}
h_d = \text{LayerNorm}(\text{GELU}(W_c[e_l;e_s])),
\end{equation}
where $e_l$ and $e_s$ are educational stage and subject embeddings. $W_c $ projects to joint space. The output $h_d $ is the instructional information feature which we need.

Despite the integration of textual transcripts, audio tracks, and video clips in existing multimodal models, these approaches often fail to effectively analyze teachers' emotions due to the obscure nature of their emotional expressions. By contrast, we convert the video clips to textual descriptions, which can alleviate the interference of a large amount of emotionally irrelevant content in the original video clips. Furthermore, we introduce instructional information as a supplementary modality to enrich the contextual understanding of the teaching environment, thereby improving the overall efficacy of the emotional analysis.

\subsection{Asymmetric Cross-Modal Interaction Module}
In order to accurately fuse the features from different modalities, we design the asymmetric cross-modal fusion module with an audio-centric strategy.

For text modality, we first place the learnable classification token (CLS) at the first location of its input sequence. Then, we use the text modality as a query and actively retrieve relevant information from the key-value pairs of the audio modality\cite{NIPS2017_3f5ee243} as follows:
\begin{equation}
h_t^{l+1} = \text{CrossAtt}(h_t^l, h_a^l) = \text{Softmax}\left(\frac{Q_t K_a^T}{\sqrt{d_k}}\right)V_a ,
\end{equation}
where $Q_t $ and $K_a$, $V_a $ are the query, key and value matrices. $h^l_t $ represents the $l$-th layer features of text modality. $h^l_a $ denotes audio features, and $d_k$ is the dimension scaling factor.
The resulting text modality is then enhanced by self-attention:
\begin{equation}
h_t^{(l)} = \text{SelfAtt}(h_t^{(l-1)}, h_t^{(l-1)}, h_t^{(l-1)}) + h_t^{(l-1)},
\end{equation}
where $\text{SelfAtt}$ denotes multi-head self-attention within each modality. $h^{(l)}_t$ represents features of text modality at layer $l$. $h^{(l-1)}_t$ preserves original features.
Finally, deeper feature representations are further extracted through feedforward neural networks as follows:
\begin{equation}
h_t^{(l)} = \text{FFN}(h_t^{(l)}) + h_t^{(l)},
\end{equation}
where $\text{FFN}$ is a two-layer perceptron with GELU activation. Applied position-wise to each token in $h^{(l)}_t$. 
For the video feature representation $h_t^{(l)}$ and the instructional information feature representation $h_d^{(l)}$, we employ the same process as text modality, and the detailed description is shown in Appendix C.

For audio modality, we also place CLS at the first location of its input sequence. Then, we use the audio modality as a query and actively retrieve relevant information from the key-value pairs of the text modality as follows:
\begin{equation}
h_a^{l+1} = \text{CrossAtt}(h_a^l, h_t^l) = \text{Softmax}\left(\frac{Q_a K_t^T}{\sqrt{d_k}}\right)V_t,
\end{equation}
where $Q_a $ and $K_t$, $V_t $ are the query, key and value matrices respectively. $h^l_a $ represents the $l$-th layer features of audio modality. $h^l_t $ denotes text features. $d_k$ is the dimension scaling factor.
The resulting audio modality is then enhanced by self-attention:
\begin{equation}
h_a^{(l)} = \text{SelfAtt}(h_a^{(l-1)}, h_a^{(l-1)}, h_a^{(l-1)}) + h_a^{(l-1)},
\end{equation}
where $\text{SelfAtt}$ denotes multi-head self-attention within each modality. $h^{(l)}_a$ represents features of audio modality at layer $l$. $h^{(l-1)}_a$ preserves original features.
Finally, deeper feature representations are further extracted through feedforward neural networks as follows:
\begin{equation}
h_a^{(l)} = \text{FFN}(h_a^{(l)}) + h_a^{(l)},
\end{equation}
where $\text{FFN}$ is a two-layer perceptron with GELU activation.Applied position-wise to each token in $h^{(l)}_a$. It is worth mentioning that in the cross-attention stage, unlike the other three modes that search related information from the key-value pairs of the audio mode, the audio modes retrieve relevant information from the key-value pairs of the text mode to achieve complementarity of the information. And we stack multiple layers of the above structures to achieve deep iterative fusion.

\subsection{ Hierarchical Feature Fusion Module}

To alleviate information redundancy of different modalities, we design the hierarchical feature fusion module.
First, the gated fusion mechanism is used to dynamically integrate the three-modal features. The CLS features of text, audio and video are spliced and inputted to the fully connected network, and normalized modal weights are generated through Softmax:
\begin{equation}
g_t, g_a, g_v = \text{Softmax}(\text{MLP}([h_t \| h_a \| h_v])), \\
\end{equation}
where $g_t,g_a,g_v $ are dynamic weights generated by softmax-normalized gates.$||$ denotes concatenation operation.$\text{MLP}$ projects the concatenated features to weight space.
Then, the feature representation after the fusion process of three modalities is obtained as follows:
\begin{equation}
h_{\text{fused}} = g_t \cdot h_t + g_a \cdot h_a + g_v \cdot h_v,
\end{equation}
where $h_{\text{fused}}$is a feature vector after the fusion of three modalities' representations. 
Finally, the gating fusion result are spliced with instructional information
feature as follows:
\begin{equation}
h_{\text{final}} = \text{MLP}([h_{\text{fused}} \| h_d]),
\end{equation}
where $h_d $ is the instructional information feature. $h_{\text{final}}$ is a feature after the fusion of all modalities.

The hierarchical fusion module preserves the dynamic traits of gated fusion while integrating semantic constraints of the teaching context, enabling the model to capture modal interactions and instructional information features simultaneously. Besides, it achieves a balance between modal complementarity and contextual relevance,  which effectively addresses the multiple requirements of emotion recognition in educational scenarios.

\subsection{Emotion Classification Module}
After obtaining the fused feature embeddings, we design an emotion classification module to get the final prediction results. In this module,
we employ a current emotion prediction classifier \cite{devlin2019bert}, 
and the prediction process is formulated as follows:
\begin{equation}
 p = \text{classifier}( h_{\text{final}}),
\end{equation}
where $p$ is the probability distribution of eight types of emotions. For classification tasks, we use standard cross entropy loss, and the specific formula is provided in the Appendix C.
\section{Experiment}
\subsection{Experimental Setup}
\textbf{Dataset and Evaluation Metrics.}
We used the T-MED dataset  to perform experiments, which is divided into 80\% as the training set, 10\% as the verification set and 10\% as the test set. We select weighted accuracy (WA) and weighted F1 values(W-F1) as metrics, which can more effectively deal with the problem of uneven distribution of emotional categories in T-MED data. And the detailed settings are provided in the Appendix D.

\textbf{Baseline Models.}
As there are few open-source multimodal teacher sentiment analysis methods, we selected nine general multimodal sentiment analysis methods as baseline methods.
The nine baseline methods are: Tensor Fusion Network (TFN)\cite{zadeh-etal-2017-tensor},  LMF\cite{liu-etal-2018-efficient-low}, Memory Fusion Network (MFN\cite{zadeh2018mfn}); Transformer-based MulT\cite{tsai-etal-2019-multimodal}, MISA\cite{hazarika_misa_2020}, ConFEDE\cite{yang-etal-2023-confede}, Semi-IIN\cite{lin2025semi}, MPLMM \cite{guo_multimodal_2024}, and MFMB\_Net\cite{MFMB-Net}. The MFMB-Net model adopts a multi-branch structure integrated feature fusion strategy to comprehensively capture the complexity of multimodal data.
And detailed descriptions of other baseline models are provided in the Appendix D.
All baseline models are implemented using the original authors' open-source code and meticulously reproduced under a unified experimental environment to ensure fair comparison.

\subsection{ Performance Comparison on T-MED}
\begin{table}[t]
\centering
\resizebox{\columnwidth}{!}{  %
\fontsize{16}{18}\selectfont 
\begin{tabular}{l|rrrr|cc}
\toprule
\textbf{Model} & \textbf{T}  & \textbf{A} &\textbf{V}  &\textbf{I} & \textbf{WA(\%)} & \textbf{W-F1(\%)} \\
\midrule
TFN (2017) &  \ding{51} & \ding{51} &\ding{51} & \ding{55} &75.23 & 74.15 \\
MFN (2018)    & \ding{51} & \ding{51} & \ding{51} &\ding{55} &77.84 & 76.71 \\
LMF(2018)  &  \ding{51}  &  \ding{51}  &\ding{51} & \ding{55} &76.72  & 75.83 \\
MuLT (2019)      &   \ding{51} & \ding{51}   & \ding{51} &\ding{55} &79.20  & 78.54 \\
MISA (2020)  &   \ding{51} & \ding{51}  & \ding{51} &\ding{55} &78.93  &78.26 \\
ConFEDE(2023) & \ding{51} & \ding{51} &\ding{51} &\ding{55} & 80.19 & 79.36 \\
MPLMM (2024)  &  \ding{51} &  \ding{51} & \ding{51} &\ding{55} &79.92   & 79.21 \\
Semi-IIN (2025)    & \ding{51} & \ding{51} &\ding{51} &\ding{55} & 80.09  & 79.53 \\
MFMB\_Net (2025)      & \ding{51}   & \ding{51}    & \ding{51} &\ding{55} &\underline{80.91}  & \underline{79.81}\\ \hline
AAM-TSA   &   \ding{51}&  \ding{51}& \ding{51} & \ding{51} &\textbf{86.84} &  \textbf{86.37} \\
\bottomrule
\end{tabular}}
\caption{Comparison results on T-MED.}
\label{4}
\end{table}

Compared with all representative multimodal emotion recognition models, 
the proposed AAM-TSA achieves significant improvements in WA and W-F1 values as shown in Table \ref{4}. The T means that methods using text features, and the A means that methods use audio features. The V means that methods use video features, and the means that methods use features of instructional information.
Compared with the current optimal baseline MFMB\_Net, the accuracy rate is increased by 5.93\%, and W-F1 is increased by 6.56\%.
We attribute it to our audio-centric design. General models often regard text modalities as core modalities rather than audio. For example, the ConFEDE architecture uses text as an anchor for comparison learning, which makes it suitable for teachers' sentiment analysis tasks. Although MPLMM also introduced prompt engineering, the MPLMM  aimed to deal with the problem of missing modalities rather than generating video description text.

\subsection{Fine-grained Sentiment Analysis}

\begin{table}[t]
\centering

\resizebox{\columnwidth}{!}{  %
\fontsize{18}{20}\selectfont 
\begin{tabular}{cccccc}
\toprule
\textbf{Model} & \textbf{Expectation} & \textbf{Neutral} & \textbf{Surprise} & \textbf{Anger}& \textbf{Joy} \\
\midrule
MFMB\_Net & 89.59 & 92.26 & 56.86 & 42.55 & 65.49\\
AAM-TSA     & 91.16 & 95.62 & 69.09 & 64.75& 81.68\\
\cmidrule(lr){1-6} 
\textbf{Model}&\textbf{Enthusiasm}& \textbf{Patience}& \textbf{Sadness}& \textbf{M-F1}& \textbf{W-F1} \\
\midrule
MFMB\_Net & 47.44 & 68.93 & 47.76 & 63.86 &79.81\\
AAM-TSA    & 60.00 & 75.76 & 55.00 & 74.13& 86.37\\
\bottomrule
\end{tabular}}
\caption{ The Results of Fine-grained sentiment analysis}
\label{5}
\end{table}

In order to evaluate the performance more carefully, we compare the proposed AAM-TSA with an SOTA baseline MFMB\_Net on eight emotions, as shown in Table \ref{5}
For five general emotions, AAM-TSA shows improvements in categories of neutral, surprise, anger, sadness  and joy.
Anger typically appears as a teaching management style rather than overt or intense anger. Its identification relies more on phonological and rhythmic features, such as tone and intensity, rather than explicit words that convey anger.
For three instructional information emotions,  AAM-TSA also shows improvements in categories of patience, enthusiasm, and expectation.
Overall, the AAM-TSA improves 10.27\% and 6.56\% in M-F1 and W-F1, respectively. Therefore, we believe that our model significantly improves the recognition ability of various emotions by integrating domain knowledge, semantic visual cues, and enhanced audio analysis.

\begin{figure}[t]
    \centering
    \includegraphics[width=0.45\textwidth]{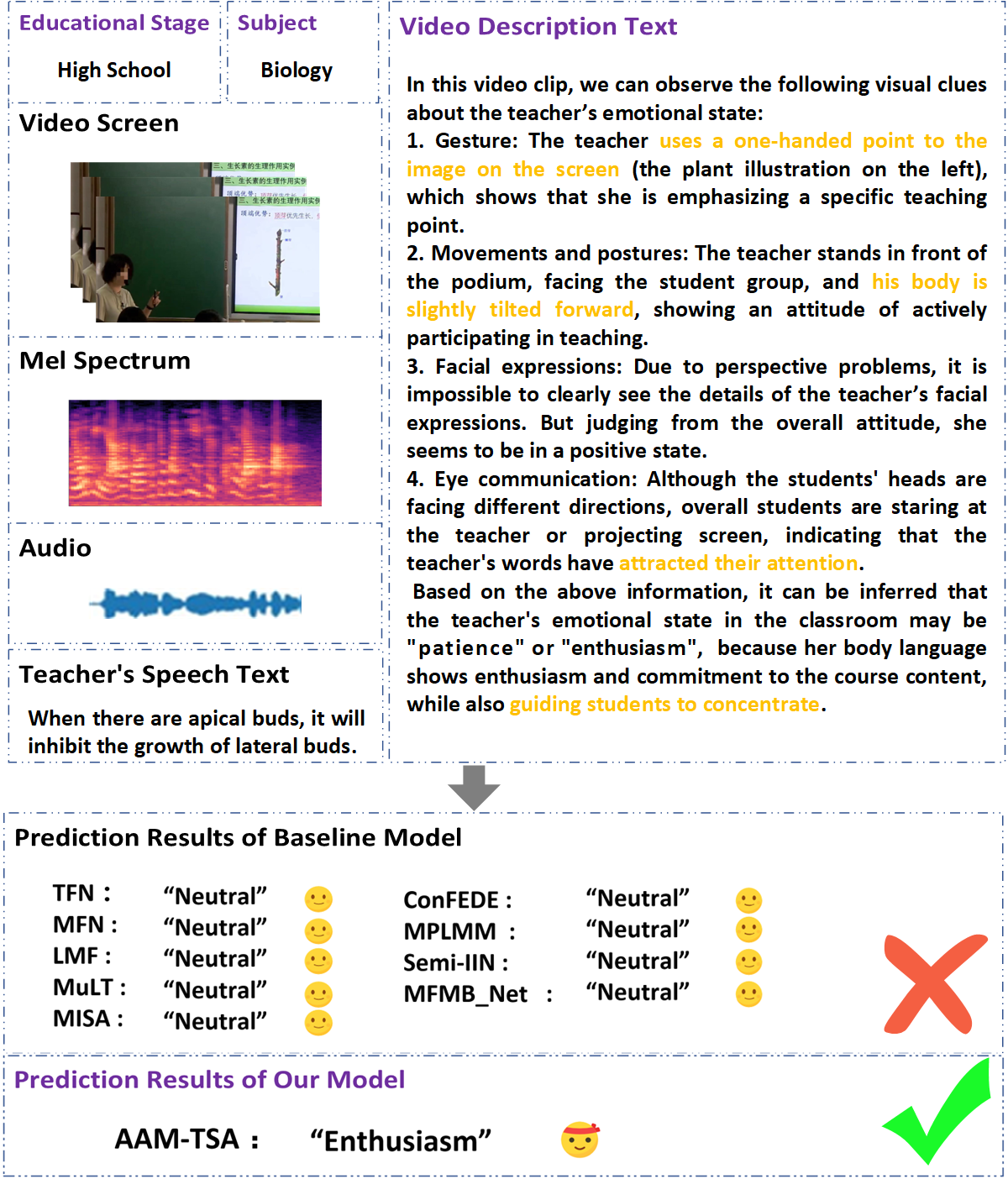}
    \caption{Analysis of predictive case.}
    \label{3}
\end{figure}
\subsection{Case Study: Teacher Sentiment Analysis in a Biology Classroom From High School}
We chose a general case to show the decision differences between our AAM-TSA and the baseline models. For all baseline models, their inputs are text, audio, and video, while our model replaces the input video with video description text and additional information for the subject and stage.
The instruction information is high school(stage)  biology (subject).
The description text, teacher's directive gestures and body forward teacher's directive gestures and body forward, furthermore reflects the teacher's positive emotional state when explaining key concepts. Combined with the characteristics of the fast speech speed and mid-frequency energy concentration of the dominant modal audio characteristics, AAM-TSA can be determined that the teacher's emotion at this time is enthusiastic, while all baseline models are misjudged as neutral, as shown in Figure \ref{3}.
In this case, text is a basic concept and has no emotional tendency. When the baseline model has no modal weight and additional information, it is easier for the baseline model to predict the case as neutral because neutrality is the majority class sample.

\subsection{The Comparasions Among Different Modalities}
To quantitatively evaluate the effectiveness of different modalities, we conduct systematic experiments on different combinations of modalities. 
We design seven varaiants and the experimental results are shown in Table \ref{6}.
The RV means that methods use raw video features directly, and the VD means that methods use features of video descriptions.

\begin{table}[t]
\centering
\resizebox{\columnwidth}{!}{  %
\fontsize{10}{12}\selectfont 
\begin{tabular}{c|cccc|cc}
\toprule
\textbf{Variants} & \textbf{T}  & \textbf{A} &\textbf{RV}  &\textbf{VD} & \textbf{WA(\%)} & \textbf{W-F1(\%)} \\
\midrule
Var-T &   \ding{51} & \ding{55} &\ding{55} & \ding{55}  & 69.41 & 64.08  \\
Var-A &   \ding{55} & \ding{51} &\ding{55} & \ding{55}  & 77.29 & 75.91 \\ 
Var-VD &   \ding{55} & \ding{55} &\ding{55} & \ding{51}      & 49.23 & 33.29 \\
\hline
Var-TA & \ding{51} & \ding{51} &\ding{55} & \ding{55}    & 85.57  & 84.74 \\
Var-TARV &\ding{51} & \ding{51} &\ding{51} & \ding{55}    & 84.97  & 84.08 \\
Var-AVD & \ding{55} & \ding{51} &\ding{55} & \ding{51}     & 78.22  & 77.69 \\
Var-TAVD &\ding{51} & \ding{51} &\ding{55} & \ding{51} & 85.97 & 85.10 \\

\bottomrule
\end{tabular}}
\caption{The Comparasions Among Different Modalities}
\label{6}
\end{table}

\textbf{The Reason for Employing Audio-centric Strategy.}
We compare the performance among audio single-mode Var-A, text single-mode single-mode Var-T, and video description Var-VD. The experimental results show that the Var-A is superior to the Var-T, with the gap between the two being 7.88\% in WA and 11.83\% in W-F1. The Var-VD  exhibits significantly poorer performance than the Var-T and the Var-A in both metrics.
This discovery intuitively confirms that the core carrier of teachers' emotional expression is audio information rather than text content.

\textbf{The Damage of Using Raw Video Features.}
We use Var-TA to denote the model using text and audio, 
and use Var-TARV to denote the model using video features directly.
Compared with the  Var-TA model, the Var-TARV  model shows a decrease of 0.60\% in terms of WA and 0.66\%  in terms of W-F1.
Furthermore, we use the proposed video description module to process the video clips and obtain two variant model denoted as Var-AVD Var-TAVD. 
The Var-AVD  exhibits significantly poorer performance than the Var-TA in both metrics.
Compared to Var-TARV The Var-TAVD model shows an increase by 0.40\% interms of WA and by 0.36\%  in terms of W-F1.
This shows that unprocessed original video features contain a large amount of emotion-independent noise, which will damage model performance.

\begin{table}[t]
\centering
\resizebox{\columnwidth}{!}{  %
\fontsize{10}{12}\selectfont 
\begin{tabular}{c|ccc|cc}
\toprule
\textbf{Varients} & \textbf{FE}  & \textbf{CI} &\textbf{HF}   & \textbf{WA(\%)} & \textbf{W-F1(\%)} \\
\midrule
Var-CIHF &  \ding{55} &\ding{51} &\ding{51} & 85.97 & 85.10 \\
Var-FEHF &  \ding{51} &\ding{55} &\ding{51}      & 86.11  & 85.43 \\

Var-FEAC &  \ding{51} &\ding{51} &\ding{55}     & 85.64   & 85.30\\
\hline
AAM-TSA &  \ding{51} &\ding{51} &\ding{51}   & 86.84 & 86.37 \\

\bottomrule
\end{tabular}}
\caption{The Ablation Study of Different Modules}
\label{7}
\end{table}

\subsection{The Ablation Study of Different Modules
}

To quantitatively evaluate the effectiveness of different modules in the AAM-TSA architecture and their contribution to final performance, we conducted systematic ablation experiments.
 We design three variants and the experimental results are shown in Table \ref{7}.
The FE means that variant models keep the multimodal feature extraction module, and CI
means that variant models keep the asymmetric cross-modal interaction module. The HF means that variant models keep the hierachical feature fusion module, and all the variant models keeps the emotion classification module

First, we construct a variant model denoted as Var-CIHF by removing the instructional information feature embedding module. 
Compared with the AAM-TSA,  the WA and W-F1 decrease 0.87\% and 1.27\%, respectively.
Thus, the introduction of school stage and subject as instructional information is crucial to understanding the situational dependence of teachers' emotional expression.

Second, we construct a variant model denoted as Var-FEHF by removing the asymmetric cross-modal attention interaction module. 
Compared with the AAM-TSA,  the WA and W-F1 decrease 0.73\% and 0.94\%.
In the teacher's sentiment analysis, the audio information carries the most significant emotional signals, allowing visual features to align the audio one-way (rather than two-way interaction), and allowing text and audio to complement each other in both directions. 
In addition, this module also allows instructional information features to actively query audio information, which can guide information flow and fusion effectively.

Fianlly, we construct a variant model denoted as Var-FEAC by removing the hierarchical feature fusion module. Compared with the AAM-TSA,  the WA and W-F1 decrease 1.2\% and 1.07\%.
This indicates that hierarchical feature fusion module can effectively utilize instructional information features and is critical to dealing with the importance differences and redundancy of multimodal information.
And more experiments and analyses that demonstrate the superiority of our design are provided in the Appendix E.

\section{Conclusion}
In this work, we address the critical gap in teacher sentiment analysis by systematically advancing both dataset construction and model development.
Firstly, we collect T-MED, the first large-scale multimodal dataset tailored for teacher sentiment analysis, through an efficient human-machine collaborative annotation framework.
Secondly, we propose the AAM-TSA model, a novel asymmetric attention-based framework designed to address the performative nature of teacher emotional expressions. By leveraging an audio-centric asymmetric attention mechanism, AAM-TSA enables differentiated cross-modal interaction, allowing effective fusion of heterogeneous features while preserving modality-specific traits. 
Extensive experimental results on T-MED demonstrate that AAM-TSA outperforms nine multimodal sentiment analysis baselines by a significant margin in terms of WA and W-F1.
This work not only provides the rigorous T-MED benchmark dataset to facilitate future research but also offers a specialized AAM-TSA model that advances the understanding of multimodal emotional expression in educational scenarios. 
By integrating pedagogical principles with technological advancements, our research paves the way for intelligent teaching tools, teacher development programs, and educational intervention systems, fostering improved teaching effectiveness and enriched student learning experiences.

\section{Acknowledgments}
This work was funded by the National Natural Science Foundation of China (Nos. 62567005 and 62406127), and Natural Science Foundation of Inner Mongolia Autonomous Region of China (No. 2025MS06004).
\FloatBarrier  

\bibliography{main}

\clearpage
\section*{Appendix}
\ifdefined\aaaianonymous
\begin{center}\textbf{Anonymous submission}\end{center}
\fi

\appendix
\setcounter{secnumdepth}{2}
\renewcommand{\thesection}{\Alph{section}.}
\renewcommand{\thesubsection}{\Alph{section}.\arabic{subsection}}

\setcounter{equation}{0}
\renewcommand{\theequation}{\arabic{equation}}
\setcounter{table}{0}
\renewcommand{\thetable}{\arabic{table}}

\section{Dataset Details}
We finally constructed a high-quality multimodal teacher classroom sentiment analysis dataset T-MED. This T-MED contains more than 200 speakers (teachers) with a total duration of more than 17 hours.

This dataset contains 14,938 sample data. Each record in the dataset is identified and positioned through the core metadata fields: the Level and Name fields uniquely identify the sample and map to the corresponding audio and video files; the Source fields indicate the class and subject to which they belong; the Start Time and End Time fields mark the time position of the fragment in the original classroom; the Duration field records the fragment duration; the Text field stores transcript text; and the Label field stores labeled emotional categories.

The emotional categories of the data set are distributed as follows: neutral (7,318 items, accounting for 49.0\%), enthusiasm (834 items, accounting for 5.6\%), joy (1,619 items, accounting for 10.8\%), patience (916 items, accounting for 6.1\%), anger (821 items, accounting for 5.5\%), expectation (2,493 items, accounting for 16.7\%), surprise (507 items, accounting for 3.4\%), sadness (430 items, accounting for 2.9\%) .

\section{Prompt Details}
In order to effectively extract the teacher’s real emotional clues in a real classroom context, we deployed the Qwen2.5-VL-7B-Instruct model locally and designed the following prompt:

\begin{quote}
Please play the role of visual sentiment analysis expert. Analyze the provided videos to identify visual cues related to the teacher’s emotional state. The emotions that teachers may have include: neutral, anger, joy, sadness, surprise, enthusiasm, patience and expectant. For teachers appearing in the video, emotional clues are given from gestures, movements, eye contact, facial expressions and other behaviors that may express emotions. If the teacher is not seen, describe the student’s behavior and interaction with the teacher.
\end{quote}

\section{Other Formulas}

\subsection{Video Modality Processing}
For video modality, we first place the learnable classification token (CLS) at the first location of its input sequence. Then we use the video modality as a query and actively retrieve relevant information from the key-value pairs of the audio modality as follows:
\begin{equation}
h_v^{l+1}=\mathrm{CrossAtt}(h_v^{l},h_a^{l})
=\mathrm{Softmax}\!\left(\frac{Q_vK_a^{T}}{\sqrt{d_k}}\right)V_a, \tag{1}
\end{equation}
where $Q_v$ and $K_a, V_a$ are the query, key and value matrices. $h_v^{l}$ represents the $l$-th layer features of video modality. $h_a^{l}$ denotes audio features. $d_k$ is the dimension scaling factor. The resulting video modality is then enhanced by self-attention:
\begin{equation}
h_v^{(l)}=\mathrm{SelfAtt}(h_v^{(l-1)},h_v^{(l-1)},h_v^{(l-1)})+h_v^{(l-1)}, \tag{2}
\end{equation}
where SelfAtt denotes multi-head self-attention within each modality. $h_v^{(l)}$ represents features of video modality at layer $l$. $h_v^{(l-1)}$ preserves original features. Finally, deeper feature representations are further extracted through feedforward neural networks:
\begin{equation}
h_v^{(l)}=\mathrm{FFN}(h_v^{(l)})+h_v^{(l)}, \tag{3}
\end{equation}
where FFN is a two-layer perceptron with GELU activation. Applied position-wise to each token in $h_v^{(l)}$.

\subsection{Instructional Information Processing}
For instructional information modality, we also place CLS at the first location of its input sequence. Then we use the instructional information modality as a query and actively retrieve relevant information from the key-value pairs of the audio modality as follow:
\begin{equation}
h_d^{l+1}=\mathrm{CrossAtt}(h_d^{l},h_a^{l})
=\mathrm{Softmax}\!\left(\frac{Q_dK_a^{T}}{\sqrt{d_k}}\right)V_a, \tag{4}
\end{equation}
where $Q_d$ and $K_a, V_a$ are the query, key and value matrices. $h_d^{l}$ represents the $l$-th layer features of instructional information modality. $h_a^{l}$ denotes audio features; $d_k$ is the dimension scaling factor. The resulting text modality is then enhanced by self-attention:
\begin{equation}
h_d^{(l)}=\mathrm{SelfAtt}(h_d^{(l-1)},h_d^{(l-1)},h_d^{(l-1)})+h_d^{(l-1)}, \tag{5}
\end{equation}
where SelfAtt denotes multi-head self-attention within each modality. $h_d^{(l)}$ represents features of domain modality at layer $l$. $h_d^{(l-1)}$ preserves original features. Finally, deeper feature representations are further extracted through feedforward neural networks:
\begin{equation}
h_d^{(l)}=\mathrm{FFN}(h_d^{(l)})+h_d^{(l)}, \tag{6}
\end{equation}
where FFN is a two-layer perceptron with GELU activation. Applied position-wise to each token in $h_d^{(l)}$.

\subsection{Loss Function}
For classification tasks, we use standard cross entropy loss as follows:
\begin{equation}
\mathcal{L}(y,z)=-\sum_{i=1}^{N}\log\left(\frac{\exp(z_{i,y_i})}{\sum_{j=1}^{n}\exp(z_{i,j})}\right), \tag{7}
\end{equation}
where $y$ is true class labels (ground truth); $z$ is predicted logits from model; $N$: Number of samples in batch; $y_i$ is the true class index for sample $i$; $z_{i,j}$ is the logit value for sample $i$ class $j$.

\section{More Experiment Content}

\subsection{Detailed Settings}
All experiments were performed on a single NVIDIA L20 GPU (48GB) and an Adamw optimizer was used. To ensure the repeatability of the experiment, we set seed to 0. The basic learning rate is set to 1e-5, the batch size is 16, the rounds are set to 30, and the attention module stacking number is 3. The training process introduced an early stop mechanism with a patience value 5 and a five-fold cross-validation was used.

\subsection{Detailed Descriptions of Other Baseline Models}
TFN adopts tensor external product fusion technology to achieve high-order cross-modal interaction for the first time, significantly improving emotional accuracy, but excessive computing overhead becomes a bottleneck. MFN innovatively combines multi-gated attention and cyclic memory units to overcome the problem of long-term dependence modeling. LMF effectively improves computational efficiency while maintaining performance flat through tensor low-rank decomposition technology.

MulT introduced directional cross-modal attention technology to break through the modal misalignment barrier. MISA successfully separates emotionally related features from modal noise through adversarial learning and feature decoupling technology. ConFEDE fusion feature decoupling and contrast learning technology to enhance the intra-class tightness of emotional features.

MPLMM designs a multimodal prompt learning framework, and effectively deals with the problem of modal missing by generating prompts, missing signal prompts, and missing type prompts triple technology. Semi-IIN is the first to create mask attention and self-training technology.

\begin{table}[t]
\centering
\begin{tabular}{l|cccc|cc}
\toprule
Variants & T & A & RV & VD & WA(\%) & W-F1(\%) \\
\midrule
Var-TARVI & \ding{51} & \ding{51} & \ding{51} & \ding{55} & 85.84 & 85.22 \\
AAM-TSA   & \ding{51} & \ding{51} & \ding{55} & \ding{51} & 86.84 & 86.37 \\
\bottomrule
\end{tabular}
\caption{The Comparasions Among Different Modalities}
\label{tab:modalities_app}
\end{table}

\begin{table}[t]
\centering
\begin{tabular}{l|ccc|cc}
\toprule
Varients & JT & FE & CI & WA(\%) & W-F1(\%) \\
\midrule
Var-FECI & \ding{55} & \ding{51} & \ding{51} & 83.30 & 82.85 \\
Var-JI   & \ding{51} & \ding{55} & \ding{55} & 85.44 & 84.52 \\
Var-JICI & \ding{51} & \ding{55} & \ding{51} & 85.97 & 85.10 \\
AAM-TSA  & \ding{51} & \ding{51} & \ding{51} & 86.84 & 86.37 \\
\bottomrule
\end{tabular}
\caption{The Ablation Study of Different Modules}
\label{tab:ablation_app}
\end{table}

\section{Other Experiments and Analysis}

\subsection{The Comparasions Among Different Modalities}
As shown in the table~\ref{tab:modalities_app}, in order to further verify the role of the video description module, we designed a variant called Var-TARVI, which was based on the complete model and replaced the video description with the original video. The experimental results show that after replacing it with the original video feature, WA decreased by 1.00\% and W-F1 decreased by 1.15\%.

\subsection{The Ablation Study of Different Modules}
To further evaluate the effectiveness of partial designs in the AAM-TSA architecture, additional ablation experiments were performed. We designed two variants, and the experimental results are shown in Table~\ref{tab:ablation_app}. JT means that the variant model retains RoBERTa and HuBERT will still participate in joint training.

First, we construct a variant model represented as Var-FECI by removing joint training. Compared with AAM-TSA, WA and W-F1 decreased by 3.54\% and 3.52\%, respectively. Therefore, joint training is crucial in real classroom scenarios. It is worth noting that our video description text cannot participate in joint training at this time, which is equivalent to weakening our advantages (the original video features themselves are difficult to participate in joint training). However, the performance of AAM-TSA is still far superior to them relative to all baseline models, which further illustrates the effectiveness of our design.

Second, based on the Var-JICI variant, we further deleted the asymmetric cross-modal attention interaction module to build a variant model represented by Var-JI. Compared with Var-JICI, WA and W-F1 decreased by 0.53\% and 0.58\%, respectively. This shows that even in the case of traditional three modes, it is necessary to design an asymmetric attention mechanism with audio as the core, and proposes an effective solution for teacher sentiment analysis.

\end{document}